\pdfoutput=1
\documentclass[11pt,a4paper]{article}

\usepackage[acceptedWithA]{styles/tacl2021v1}

\usepackage{times}
\usepackage{latexsym}

\usepackage[T1]{fontenc}

\usepackage[utf8]{inputenc}

\usepackage{microtype}

\usepackage{graphicx}
\usepackage{tabularx}
\usepackage{array}
\usepackage{booktabs}
\usepackage{algpseudocode}
\usepackage{algorithm}
\usepackage{amsmath}
\usepackage{enumitem}
\usepackage[table]{xcolor}
\usepackage{soul}

\newcommand\BibTeX{B\textsc{ib}\BibTeX}
\newcommand{\highlight}{\cellcolor{yellow!45}}

\usepackage{xspace,mfirstuc,tabulary}

\newif\iftaclinstructions
\taclinstructionsfalse 
\iftaclinstructions
\renewcommand{\confidential}{}
\renewcommand{\anonsubtext}{(No author info supplied here, for consistency with
TACL-submission anonymization requirements)}
\newcommand{\instr}
\fi

\iftaclpubformat 

\else

\fi

\title{The Bias is in the Details: An Assessment of Cognitive Bias in LLMs}
\author{
  \textbf{R. Alexander Knipper}\footnotemark[1]\quad
  \textbf{Charles S. Knipper}\footnotemark[1]\quad
  \textbf{Kaiqi Zhang}\footnotemark[2] \\
  \textbf{Valerie Sims}\footnotemark[2]\quad
  \textbf{Clint Bowers}\footnotemark[2]\quad
  \textbf{Santu Karmaker}\footnotemark[1] \\
  Bridge-AI Lab@UCF, Department of Computer Science\footnotemark[1] \\
  Department of Psychology\footnotemark[2] \\
  University of Central Florida, USA \\
  \texttt{\{alexknipper, charlie.knipper, kaiqi.zhang\}@ucf.edu}\\
  \texttt{\{Valerie.Sims, Clint.Bowers, santu\}@ucf.edu} \\}
\date{June 2025}
\begin{document}

\maketitle

\begin{abstract}

As Large Language Models (LLMs) are increasingly embedded in real-world decision-making processes, it becomes crucial to examine the extent to which they exhibit \textit{cognitive biases}. Extensively studied in the field of psychology, cognitive biases appear as systematic distortions commonly observed in human judgments. This paper presents a large-scale evaluation of \textbf{eight} well-established cognitive biases across \textbf{45 LLMs}, analyzing over \textbf{2.8 million} LLM responses generated through controlled prompt variations. To achieve this, we introduce a novel evaluation framework based on multiple-choice tasks, hand-curate a dataset of 220 decision scenarios targeting fundamental cognitive biases in collaboration with psychologists, and propose a scalable approach for generating diverse prompts from human-authored scenario templates. 
Our analysis shows that LLMs exhibit bias-consistent behavior in \textbf{17.8}-\textbf{57.3\%} of instances
across a range of judgment and decision-making contexts targeting anchoring, availability, confirmation, framing, interpretation, overattribution, prospect theory, and representativeness biases. 
%
We find that both model size and prompt specificity play a \emph{significant} role on bias susceptibility as follows: larger size (>32B parameters) can reduce bias in \textbf{39.5\%} of cases, while higher prompt detail reduces most biases by up to \textbf{14.9\%}, except in one case (Overattribution), which is exacerbated by up to \textbf{8.8\%}.

\end{abstract}


\section{Introduction}\label{sec:intro}


Intelligent decision support systems are increasingly leveraging Large Language Models (LLMs) to generate summaries \cite{laban-etal-2023-summedits, zhang2024benchmarking, song-etal-2025-learning}, analyze scenarios \cite{dagan2023dynamicplanningllm, wang-etal-2024-planning}, answer questions \cite{yiming-etal-2023-qa, dean-juan-2024-qa}, and provide recommendations across domains such as healthcare, finance, crisis response, and policy planning \cite{yangyang-etal-2024-decisionmaking, shool2025systematic}. With their ability to process vast amounts of unstructured data, LLMs have become integral to supporting complex tasks involving \emph{judgment} and \emph{decision-making}.

However, as LLMs are primarily trained on human-generated text and learn to replicate human language patterns, they also tend to reflect a range of human-like biases \cite{schramowski-2022-bias-right-and-wrong}—including cultural \cite{tao-etal-2024-cultural}, ethnic \cite{javed2025llms}, gender-based \cite{kotek-et-al-2023-gender, farlow2024gender}, attributional \cite{gandhi2023understanding}, and probabilistic biases \cite{wang2024will, ryu2024study}. These biases can distort both the content of AI-generated responses and the underlying \emph{reasoning} used to justify them.
As a result, biased LLM outputs may influence users' perceptions, judgments, and decisions,
raising serious concerns about the reliability, fairness, and rationality of LLM-based decision support systems \cite{lai2023towards}.
Among the many types of bias, comparatively less attention has been given to \emph{cognitive biases}—a class of systematic distortions that affect human reasoning, judgment, and decision-making. These biases have been studied extensively in psychology as a framework for understanding how individuals deviate from normative or rational thinking \cite{tversky1974judgment}. Cognitive biases often arise from heuristics—mental shortcuts that simplify complex decisions but can lead to flawed conclusions, misinterpretations, or illogical judgments \cite{kahneman2011thinking}. They are well-documented in psychology and known to affect perception, memory, reasoning, and behavior across a wide range of contexts \cite{FURNHAM201135, caputo2013literature}.

As LLMs are increasingly embedded in systems designed to support or simulate human judgment and decision-making, it becomes essential to evaluate the extent to which they exhibit these cognitive biases.
This inquiry is particularly urgent given the widespread deployment of LLMs in high-stakes domains such as education, hiring, law, and healthcare \cite{kasneci2023chatgpt, vrdoljak2025review}, where even subtle distortions in reasoning can propagate into significant real-world consequences \cite{choi2024llm, spatharioti2025effects}.
Consequently, understanding how cognitive biases manifest in LLMs is essential for evaluating their reliability, trustworthiness, and rationality in judgment and decision-making contexts.


In this paper, we conduct a large-scale evaluation of \textbf{eight} targeted cognitive biases across \textbf{45} LLMs, analyzing more than $\mathbf{2.8}$ million total LLM responses generated through controlled prompt variations using the TELeR prompting taxonomy \cite{teler}. To achieve this, we make the following contributions in this paper:

\vspace{-2mm}
\begin{enumerate}[leftmargin=*,itemsep=0ex,partopsep=0ex,parsep=0ex]
    
    

    \item Develop a comprehensive evaluation framework for cognitive biases in LLMs, a curated dataset of 220 psychologist-authored judgment and decision scenarios, and a scalable method for generating diverse prompt variations from expert-written templates.
    

    \item Conduct extensive experiments across $45$ LLMs to evaluate susceptibility to cognitive bias, analyzing the effects of model size, temperature, and prompt specificity.

    \item Show that LLMs exhibit bias-consistent behavior in \textbf{17.8}-\textbf{57.3\%} of judgment and decision-making situations. Both \textit{model size} and \textit{prompt specificity} play a significant role in the level of bias present, reducing bias expression by up to \textbf{39.5\%} and \textbf{14.9\%} respectively.

\end{enumerate}
\section{Related Work}\label{sec:related}






Existing research on biases in LLMs can be broadly categorized into two primary domains: (1) \emph{social biases} and (2) \emph{cognitive biases}. Of these, recent literature has predominantly focused on social biases \cite{salecha2024largelanguagemodelshumanlike, fan-etal-2024-biasalert, 2025biasedassociations-social}, particularly those related to gender \cite{wan-etal-2023-kelly, kotek-et-al-2023-gender, farlow2024gender}, ethnicity \cite{javed2025llms}, socioeconomic status \cite{Arzaghi_Carichon_Farnadi_2024, singh2024bornsilverspooninvestigating}, and other identity markers \cite{wilson2024gender, demidova2024john, tao-etal-2024-cultural}. Such biases have been extensively documented in both open-ended generation and structured tasks \cite{2021stochasticparrots}, often being attributed to imbalances in training data or reinforcement learning feedback. Several mitigation strategies have been proposed \cite{Ling_Rabbi_Wang_Yang_2025, 2025breakingbiasbuildingbridges-eval-mitigation-social}, though their effectiveness remains mixed across settings.
While social bias is not the focus of this work, its persistence underscores the broader need to understand and address bias in LLM reasoning.

Similarly, a complementary line of work has examined the presence of \emph{cognitive biases} in LLMs \cite{Binz_2023}, showing that models can replicate a wide range of human-like biases \cite{itzhak-etal-2024-instructed,malberg-etal-2025-comprehensive}.
Prior studies have typically focused on confirming specific biases in isolated settings \cite{koo-etal-2024-benchmarking, chen-cognitive-bias-threshold-priming-2024}, or explored interventions to mitigate them \cite{echterhoff-etal-2024-cognitive, yasuaki-etal-cognitive-survey-mitigation}.
However, there has been limited analysis of how \emph{prompt structure} \cite{tjuatja-etal-2024-llms}—particularly the \emph{level of detail}—influences a model's susceptibility to cognitive bias. This is a critical oversight given the variability of real-world LLM use-cases, where prompts often differ significantly in specificity and scope.





We address this gap by designing controlled prompt perturbations that systematically vary prompt detail \cite{teler}, enabling direct measurement of how bias susceptibility in LLMs changes across levels of specificity.
Our experimental framework draws from established psychological methodologies, which favor single-shot, questionnaire-style assessments commonly delivered in multiple-choice format \cite{berthet2023heuristics}.
While the cognitive sciences have identified a broad range of biases \cite{blumenthal2015cognitive, flyvbjerg2021top}, only a core subset have been shown to be both reliably inducible \cite{saposnik2016cognitive} and relevant for mitigation-focused analysis \cite{korteling2021retention}.
Accordingly, our study centers on eight canonical biases that span key cognitive heuristics and support robust evaluations under controlled variations in prompt detail.

\section{Selection of Cognitive Biases}\label{sec:background}






We identify eight distinct biases to represent a commonly observed, robust, and relevant set of cognitive biases for evaluation. Though many biases are defined by existing literature \cite{blumenthal2015cognitive, flyvbjerg2021top}, not all are necessary or relevant for a holistic assessment. Some biases may be specific effects of a broader mechanism \cite{tversky1983extensional}, not yet robustly replicated \cite{ayton2004hot}, or rooted in mechanisms theoretically irrelevant to LLMs \cite{dror2020cognitive}. Biases in these categories were therefore excluded from consideration.

The biases selected represent those that are most common in critical judgment and decision making scenarios \cite{saposnik2016cognitive}, reflecting important abilities within high-stakes domains, including: assessment of uncertain values, inference of cause and effect, attribution of intent or emotion, and probability estimation \cite{morewedge2015debiasing}.

Based on the criteria, we identify these eight biases as a holistic characterization of cognitive bias in LLMs: Anchoring, Availability, Confirmation, Framing, Interpretation, Overattribution, Prospect Theory, and Representativeness bias. We refer to each bias with the following definitions, and provide an example of how each bias might manifest in a hypothetical scenario.

\begin{figure*}[ht]
    \centering
    \includegraphics[scale = 0.425]{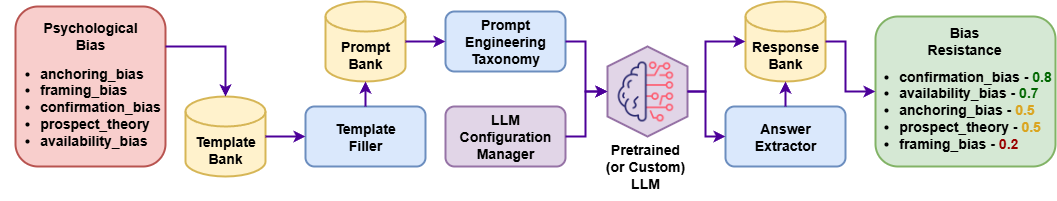}
    \caption{Psychological bias assessment architecture. Each bias is measured using a variety of prompts derived from a series of hand-crafted template queries.}
    \label{fig:architecture}
\end{figure*}
\vspace{-2pt}

\subsection{Anchoring Bias}
Anchoring bias occurs when an initial, "anchor" value becomes the basis for estimation in future judgments \cite{tversky1974judgment}.

\textbf{Example:} \emph{"You read that 130,000 people traveled on Florida interstates last year; likely mostly tourists. How many people do you think actually live in Florida?"}

A response in the hundreds of thousands reflects bias by relying too heavily on the first value mentioned, despite Florida’s population exceeding 20 million.



\subsection{Availability Bias}
Availability bias occurs when judgments of likelihood are conflated with how readily something can be imagined or remembered \cite{tversky1974judgment}.

\textbf{Example:} \emph{"While planning a trip, you recall a friend’s story about a turbulent flight that ended in an emergency landing. Do you think flying or driving is safer for your trip?"}

Choosing "driving" reflects bias by overly weighting the readily available example, despite flying being statistically safer than driving.



\subsection{Confirmation Bias}
Confirmation bias reflects the tendency to make judgments that align with, rather than disprove, existing beliefs \cite{wason1960failure}.

\textbf{Example:} \emph{"You have a fever, sore throat, and fatigue; probably the flu. The doctor notes that you might be showing signs of strep, but it's too early to tell. Do you ask for flu or strep medication?"}

Choosing flu medication reflects bias by preferring evidence for the assumed diagnosis, while ignoring contradicting evidence of other illnesses.



\subsection{Framing Bias}
Framing bias occurs when differences in the presentation of information leads to different judgments \cite{tversky1981framing}.

\textbf{Example:} \emph{"You're choosing between two surgeons. One has an 11\% mortality rate; the other, an 83\% success rate. Both are equally qualified. Who do you choose?"}

Choosing the second doctor reflects bias by favoring the positively framed statistic, despite their success rate actually being lower (83\% $<$ 89\%).



\subsection{Interpretation Bias}
Interpretation bias involves attributing positive or negative meaning to an ambiguous situation \cite{mathews1994cognitive}.

\textbf{Example:} \emph{"You enter a room full of people who begin laughing. They've noticed you, but you're not sure what they're laughing about. Do you see this as good, bad, or neither?"}

Responding "good" or "bad" reflects bias by imposing a specific meaning onto an ambiguous situation with no explicitly stated cause.



\subsection{Overattribution Bias}
Overattribution bias reflects the tendency to overly attribute behaviors to personal characteristics, rather than external factors \cite{ross1977intuitive}.

\textbf{Example:} \emph{"You're interviewing a highly qualified candidate who seems a little irritable and disengaged. Do you ask if something’s wrong, or conclude they’re a poor fit?"}

Choosing "the candidate is a poor fit" reflects bias by assuming that this behavior reflects their character, without considering situational factors.



\subsection{Prospect Theory Bias}
Prospect Theory bias involves judging a prospective loss as more significant than an equivalent prospective gain \cite{kahneman2013prospect}.

\textbf{Example:} \emph{"A game presents you with two deals, you can either take a sure 100 coins, or gamble: 80\% chance of gaining 200 coins, 20\% chance of losing 50 coins. Which do you pick?"}

Choosing the sure 100 coins reflects bias by overly avoiding a possible loss, despite the higher average return of the gamble (150 coins).



\subsection{Representativeness Bias}
Representativeness bias occurs when judgments of likelihood are conflated with how closely something follows preconceived expectations \cite{tversky1974judgment}.

\textbf{Example:} \emph{"You're introduced to two people— you know one's an electrician, the other's a therapist. One is wearing overalls and boots; the other, dress pants and a shirt. Who's the therapist?"}

Choosing the individual in dress clothes reflects bias by stereotyping their appearances to their occupations, despite equal likelihood for either to be the therapist.



\section{Methodology}\label{sec:methods}

In this section, we provide an overview of our systematic evaluation process. Figure \ref{fig:architecture} depicts the proposed pipeline, where we turn hand-crafted template scenarios into an automated evaluation for each of our selected biases.

\subsection{Prompt Generation}\label{subsec:generation}


To consistently measure whether language models exhibit cognitive biases, we first construct a diverse, realistic set of prompts designed to elicit bias-consistent responses. This requires balancing three core objectives: \textbf{(1)} generating a sufficient number of prompts for robust analysis, \textbf{(2)} introducing sufficient variation to avoid lexical or structural overfitting, and \textbf{(3)} maintaining naturalistic phrasing that mirrors real-world user interactions.



We address these objectives through a two-stage prompt generation pipeline that combines human-authored templates with automated augmentation. This process begins with a set of 20-40 manually-written \textbf{template scenarios} per bias. Each template captures the core structure of a decision-making situation, abstracted away from specific surface details. For example:

\textcolor{blue}{\emph{I am a <craftsman\_job> trying to figure out the average weight of <construction\_tool>...}}

These templates serve as the high-level scaffolding for prompt construction. We denote the full collection as $T$, the template set.

Next, we instantiate each template into multiple concrete variants using a \textbf{template filling} process. Each placeholder tag (e.g., \textit{<craftsman\_job>}) is resolved using one of two strategies:

\begin{itemize}\setlength\itemsep{0cm}
    \item \textbf{Phrases} are randomly sampled from curated slot-specific lists.
    \item \textbf{Numbers} are generated using parameterized arithmetic expressions, optionally referencing ranges and/or previously-filled values (e.g. \emph{<}$percentage1 - [50, 75] / 100$\emph{>})
\end{itemize}

This filling procedure yields a final set $S$ of fully-specified \textbf{scenario instances}, where each template produces $k = 25$ unique variations. In total, this yields $500$-$1,000$ instances per bias category, enabling broad and controlled evaluation. In the next step, we systematically adjust the level of contextual detail presented in each prompt to study how bias expression varies with prompt specificity.

\subsection{Prompt Engineering}

\begin{figure}[ht]
    \centering
    \includegraphics[scale = 0.275]{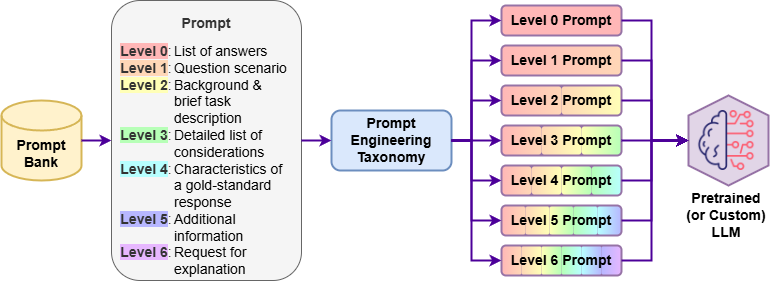}
    \caption{Prompt engineering process used to vary the level of detail being given to the LLM. Each prompt is split into five levels of detail, then sent to the LLM for evaluation in all forms.}
    \label{fig:prompt_engineering}
\end{figure}

To systematically evaluate the impact of prompt specificity on bias expression, we decompose each scenario instance into five versions with increasing levels of detail, following the TELeR prompt engineering framework \cite{teler}. This results in a set of prompts that vary in informational richness while preserving the underlying scenario structure, as shown in Figure~\ref{fig:prompt_engineering}.

We focus on TELeR Levels $1$-$5$, which progressively introduce scenario-relevant context. Lower levels present only minimal task framing, while higher levels add explanatory background, surface constraints, and clarifying information. For Level 5, we manually simulate a retrieval-augmented generation (RAG) setup by injecting additional, truthful information about the scenario. These additions are carefully curated to avoid introducing artifacts that might unintentionally suppress or amplify the bias being measured.

Applying this transformation to the set of instantiated scenarios $S$ yields our final prompt set $P$, with $P = |S| \times 5$. This results in approximately $2,500$-$5,000$ total prompts per bias, enabling controlled, multi-resolution analysis of LLM susceptibility to bias as a function of prompt specificity.

\subsection{Answer Extraction}\label{subsec:extraction}

Once prompts are submitted, we collect the model's free-form responses and extract the model's answer probabilistically. To maintain a naturalistic evaluation setting, we avoid constraining models to a fixed output format.

Given a model response $r \in R$ and a set of answer choices $A$, we compute a confidence score $S_a$ for each $a \in A$ based on two main signals:

\begin{itemize}\setlength\itemsep{0cm}
    \item \textbf{Presence}: Whether the answer is mentioned in the response, accounting for lexical variation via fuzzy string matching.
    \item \textbf{Sentiment}: The local connotation surrounding the answer mention, based on proximity to sentiment-laden terms.
\end{itemize}

To compute \textbf{presence}, we scan the generated text using a sliding window and calculate string similarity scores between candidate spans and each answer choice. Matches above a similarity threshold ($t = 0.8$) are used to estimate how explicitly an answer is referenced.

For \textbf{sentiment}, we assess the polarity of words surrounding each detected mention, measuring its proximity to both positive and negative terms. This provides a signal of endorsement or rejection, even when multiple choices are referenced.

Each answer's final score $S_a$ is computed by weighting its presence score with the surrounding sentiment context. The answer with the highest score is taken to be the model's implicit choice. This approach preserves the flexibility of unconstrained generation while enabling reliable, automated answer attribution.
\begin{figure*}[ht]
    \centering
    \includegraphics[scale = 0.45]{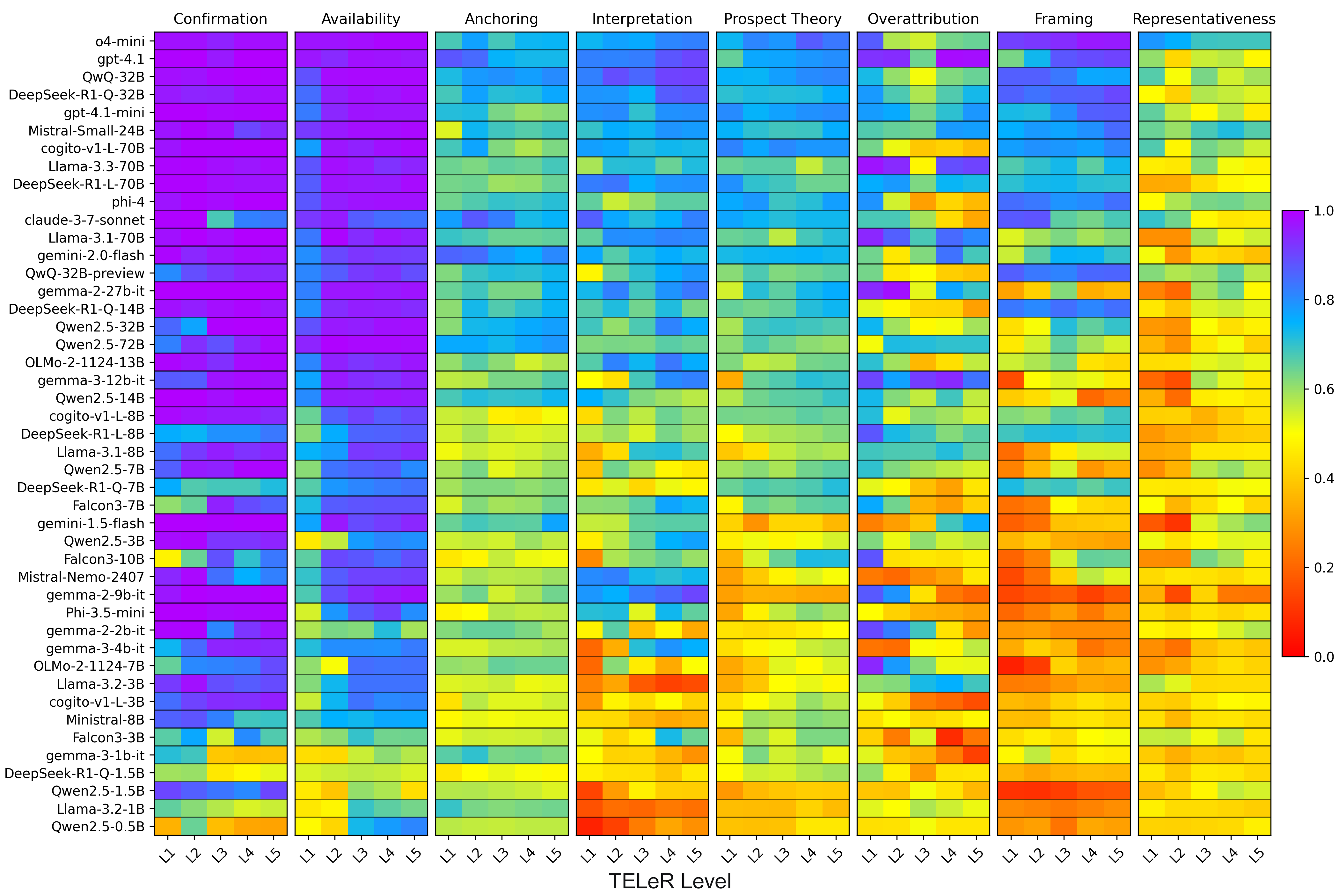}
    \vspace{-8mm}
    \caption{Heatmaps illustrating model resistance against bias. Each subplot is a distinct bias; rows indicate models tested (ordered by score) and columns denote TELeR levels (1–5). Colors reflect average bias resistance scores, where $\uparrow$ higher scores (cooler colors) indicate resistance and $\downarrow$ lower scores (warmer colors) indicate susceptibility.} 
    \label{fig:heatmap_score}
\end{figure*}
\section{Experimental Setup}\label{sec:experiments}

In this section, we detail our evaluation setup—how the evaluation data is organized, which models are tested, any important parameters, and how we define and measure a model's bias susceptibility.

\subsection{Dataset}

Our evaluation dataset consists of $220$ total scenario templates, which, after the template-filling operation, expands to $5,500$ unique instances. Following the prompt-splitting operation, the total number of unique prompts increases to $27,500$ per model, per temperature, across all biases being tested. Specifically, the dataset includes the following distribution:
\begin{itemize}[leftmargin=*,itemsep=0ex,partopsep=0ex,parsep=0ex]
    \item $5,000$ prompts for Anchoring and Representativeness biases
    \item $3,750$ prompts for Availability and Framing biases
    \item $2,500$ prompts for Confirmation, Overattribution, Prospect Theory, and Interpretation biases
\end{itemize}

\subsection{Models Examined}


We perform tests on a total of $45$ models from a wide spectrum of the following model families:
Phi 3.5 \& 4 \cite{abdin2024phi3technicalreporthighly, abdin2024phi4technicalreport},
Qwen 2.5 \& QWQ \cite{qwen2.5, qwq32b},
DeepSeek-R1 \cite{deepseekai2025deepseekr1incentivizingreasoningcapability},
Cogito v1 (Preview),
LLaMA 3 \cite{grattafiori2024llama3herdmodels},
Falcon 3 \cite{Falcon3},
OLMo 2 \cite{olmo20252olmo2furious},
Mistral \& Ministral \cite{jiang2023mistral7b},
Gemma 2 \& 3 \cite{gemmateam2024gemma2improvingopen, gemmateam2025gemma3technicalreport},
GPT 4.1 \& o4 \cite{openai2024gpt4technicalreport},
Gemini 1.5 \& 2.0 \cite{geminiteam2024geminifamilyhighlycapable, geminiteam2024gemini15unlockingmultimodal}, and
Claude 3.7 \cite{Claude3S}.

\subsection{Model Parameters}

For all models, we fix the $top\_k$ and $top\_p$ parameters at their default values of $-1$ and $1$, respectively \cite{kwon2023efficient-vllm}, ensuring that all tokens are considered during generation. We vary the temperatures across $20$ models, ranging from $0.2$ to $0.8$, while the remaining $25$ models are tested at a single temperature: $24$ models at the lowest temperature in the range ($0.2$), and one model at its lowest supported temperature ($1.0$). This results in a total of $2,887,500$ model responses evaluated across all models, TELeR levels, temperatures, and biases.



\subsection{Metrics}


Since bias is not solely determined by the correctness of a response, we classify each answer as either \emph{biased}, \emph{unbiased}, or \emph{unrelated}, independent of the correctness of the answer. Only responses marked as \emph{biased} are considered in the evaluation. The \emph{bias resistance score} is then computed as the complement of the ratio of biased responses across the total set, providing a quantitative measure of a model's tendency to produce biased output, regardless of its overall accuracy.
\section{Results}\label{sec:results}

\subsection{Bias Susceptibility Analysis}



We begin our analysis by calculating bias resistance scores for each combination of model, bias, and TELeR prompt level, producing a comprehensive matrix of scores, as shown in Tables \ref{tbl:performance_models} and \ref{tbl:performance_teler}. Overall, we find that nearly all models exhibit measurable bias-consistent behavior, with average susceptibility ranging from \textbf{17.8\%} to \textbf{57.3\%}.

We further observe two overarching trends: \textbf{(1)} larger models tend to exhibit lower susceptibility, and \textbf{(2)} prompts with greater contextual detail (i.e., higher TELeR levels) often lead to improved resistance. However, the strength of these effects varies by biases and model type, suggesting opportunities for targeted mitigation through prompt design and architectural tuning.


\begin{table*}[!htb]
    \centering
    \resizebox{\textwidth}{!}{
    \begin{tabular}{c||c|c|c|c|c|c|c|c|c}
        \hline
         Model
         & \begin{tabular}[c]{@{}c@{}} Anchoring\end{tabular}
         & \begin{tabular}[c]{@{}c@{}} Availability\end{tabular}
         & \begin{tabular}[c]{@{}c@{}} Confirmation\end{tabular}
         & \begin{tabular}[c]{@{}c@{}} Framing\end{tabular}
         & \begin{tabular}[c]{@{}c@{}} Overattribution\end{tabular}
         & \begin{tabular}[c]{@{}c@{}} Prospect Theory\end{tabular}
         & \begin{tabular}[c]{@{}c@{}} Representativeness\end{tabular}
         & \begin{tabular}[c]{@{}c@{}} Interpretation\end{tabular}
         & \begin{tabular}[c]{@{}c@{}} Average\end{tabular}\\
         \hline\hline
phi-4		& 0.680	& 0.950	& 0.989	& 0.822	& 0.486	& 0.743	& 0.592	& 0.624	& 0.736	\\
\hline
Phi-3.5-mini		& 0.536	& 0.782	& 0.991	& 0.262	& 0.383	& 0.513	& 0.428	& 0.668	& 0.570	\\
\hline
QwQ-32B-preview		& 0.695	& 0.883	& 0.898	& 0.845	& 0.475	& 0.641	& 0.602	& 0.676	& 0.714	\\
\hline
QwQ-32B		& 0.776	& 0.965	& 0.987	& 0.818	& 0.622	& 0.774	& 0.587	& \highlight\textbf{0.874}	& 0.800	\\
\hline
Qwen2.5-72B		& 0.761	& 0.980	& 0.914	& 0.558	& 0.669	& 0.626	& 0.405	& 0.638	& 0.694	\\
\hline
Qwen2.5-32B		& 0.723	& 0.952	& 0.922	& 0.602	& 0.582	& 0.664	& 0.401	& 0.706	& 0.694	\\
\hline
Qwen2.5-14B		& 0.703	& 0.929	& 0.995	& 0.365	& 0.632	& 0.645	& 0.391	& 0.646	& 0.663	\\
\hline
Qwen2.5-7B		& 0.580	& 0.799	& 0.951	& 0.358	& 0.603	& 0.617	& 0.470	& 0.509	& 0.611	\\
\hline
Qwen2.5-3B		& 0.562	& 0.682	& 0.953	& 0.348	& 0.572	& 0.502	& 0.500	& 0.640	& 0.595	\\
\hline
Qwen2.5-1.5B		& 0.559	& 0.493	& 0.861	& 0.132	& 0.418	& 0.372	& 0.472	& 0.345	& 0.456	\\
\hline
Qwen2.5-0.5B		& 0.563	& 0.645	& 0.403	& 0.296	& 0.459	& 0.410	& 0.428	& 0.211	& 0.427	\\
\hline
DeepSeek-R1-L-70B		& 0.624	& 0.949	& 0.983	& 0.717	& 0.726	& 0.693	& 0.417	& 0.799	& 0.738	\\
\hline
DeepSeek-R1-Q-32B		& 0.728	& 0.944	& 0.963	& 0.865	& 0.687	& 0.722	& 0.515	& 0.814	& 0.780	\\
\hline
DeepSeek-R1-Q-14B		& 0.690	& 0.915	& 0.974	& 0.826	& 0.435	& 0.677	& 0.489	& 0.674	& 0.710	\\
\hline
DeepSeek-R1-L-8B		& 0.551	& 0.792	& 0.783	& 0.705	& 0.714	& 0.573	& 0.357	& 0.584	& 0.632	\\
\hline
DeepSeek-R1-Q-7B		& 0.613	& 0.787	& 0.702	& 0.686	& 0.433	& 0.674	& 0.479	& 0.483	& 0.607	\\
\hline
DeepSeek-R1-Q-1.5B		& 0.493	& 0.550	& 0.532	& 0.355	& 0.455	& 0.550	& 0.438	& 0.441	& 0.477	\\
\hline
cogito-v1-L-70B		& 0.655	& 0.932	& 0.992	& 0.788	& 0.465	& 0.792	& 0.587	& 0.742	& 0.744	\\
\hline
cogito-v1-L-8B		& 0.510	& 0.835	& 0.963	& 0.642	& 0.598	& 0.641	& 0.403	& 0.574	& 0.646	\\
\hline
cogito-v1-L-3B		& 0.524	& 0.750	& 0.914	& 0.399	& 0.303	& 0.527	& 0.451	& 0.423	& 0.536	\\
\hline
Llama-3.3-70B		& 0.650	& 0.940	& 0.983	& 0.697	& 0.837	& 0.635	& 0.506	& 0.674	& 0.740	\\
\hline
Llama-3.2-3B		& 0.535	& 0.773	& 0.909	& 0.287	& 0.676	& 0.445	& 0.480	& 0.209	& 0.539	\\
\hline
Llama-3.2-1B		& 0.643	& 0.584	& 0.586	& 0.262	& 0.535	& 0.378	& 0.431	& 0.208	& 0.453	\\
\hline
Llama-3.1-70B		& 0.664	& 0.936	& 0.991	& 0.590	& 0.828	& 0.654	& 0.444	& 0.774	& 0.735	\\
\hline
Llama-3.1-8B		& 0.542	& 0.857	& 0.919	& 0.415	& 0.679	& 0.512	& 0.408	& 0.572	& 0.613	\\
\hline
Falcon3-10B		& 0.504	& 0.832	& 0.709	& 0.459	& 0.538	& 0.594	& 0.446	& 0.544	& 0.578	\\
\hline
Falcon3-7B		& 0.596	& 0.848	& 0.793	& 0.360	& 0.494	& 0.613	& 0.444	& 0.678	& 0.603	\\
\hline
Falcon3-3B		& 0.545	& 0.633	& 0.689	& 0.469	& 0.301	& 0.547	& 0.530	& 0.555	& 0.534	\\
\hline
OLMo-2-1124-13B		& 0.600	& 0.918	& 0.974	& 0.526	& 0.532	& 0.607	& 0.493	& 0.760	& 0.676	\\
\hline
OLMo-2-1124-7B		& 0.627	& 0.728	& 0.799	& 0.260	& 0.677	& 0.456	& 0.376	& 0.424	& 0.543	\\
\hline
Mistral-Small-24B		& 0.662	& 0.964	& 0.958	& 0.790	& 0.704	& 0.716	& 0.661	& 0.752	& 0.776	\\
\hline
Mistral-Nemo-2407		& 0.572	& 0.858	& 0.869	& 0.374	& 0.301	& 0.445	& 0.443	& 0.758	& 0.577	\\
\hline
Ministral-8B		& 0.511	& 0.735	& 0.788	& 0.412	& 0.460	& 0.574	& 0.425	& 0.381	& 0.536	\\
\hline
gemma-2-27b-it		& 0.667	& 0.935	& \highlight\textbf{1.000}	& 0.412	& 0.780	& 0.680	& 0.437	& 0.771	& 0.710	\\
\hline
gemma-2-9b-it		& 0.600	& 0.886	& 0.990	& 0.158	& 0.508	& 0.334	& 0.275	& 0.824	& 0.572	\\
\hline
gemma-2-2b-it		& 0.626	& 0.626	& 0.939	& 0.287	& 0.629	& 0.458	& 0.510	& 0.466	& 0.568	\\
\hline
gemma-3-12b-it		& 0.613	& 0.908	& 0.935	& 0.432	& 0.872	& 0.612	& 0.385	& 0.648	& 0.676	\\
\hline
gemma-3-4b-it		& 0.559	& 0.787	& 0.884	& 0.321	& 0.399	& 0.510	& 0.338	& 0.559	& 0.545	\\
\hline
gemma-3-1b-it		& 0.651	& 0.521	& 0.509	& 0.486	& 0.330	& 0.555	& 0.389	& 0.395	& 0.479	\\
\hline
o4-mini		& 0.721	& \highlight\textbf{0.982}	& 0.972	& \highlight\textbf{0.941}	& 0.656	& \highlight\textbf{0.806}	& \highlight\textbf{0.719}	& 0.779	& \highlight\textbf{0.822}	\\
\hline
gpt-4.1		& 0.786	& 0.963	& 0.992	& 0.810	& \highlight\textbf{0.894}	& 0.752	& 0.527	& 0.847	& 0.821	\\
\hline
gpt-4.1-mini		& 0.656	& 0.935	& 0.994	& 0.795	& 0.733	& 0.783	& 0.547	& 0.779	& 0.778	\\
\hline
gemini-2.0-flash		& \highlight\textbf{0.808}	& 0.887	& 0.972	& 0.679	& 0.648	& 0.736	& 0.406	& 0.727	& 0.733	\\
\hline
gemini-1.5-flash		& 0.683	& 0.899	& 0.998	& 0.318	& 0.480	& 0.380	& 0.403	& 0.620	& 0.598	\\
\hline
claude-3-7-sonnet		& 0.787	& 0.886	& 0.864	& 0.745	& 0.540	& 0.737	& 0.546	& 0.782	& 0.736	\\
\hline
    \end{tabular}
    }
    \caption{Average bias resistance scores for each model. Higher ($\uparrow$) scores indicate resistance, and lower ($\downarrow$) scores indicate susceptibility. Top performers are highlighted and marked in \highlight\textbf{bold}. Models suffixed with “-Q” are Qwen-distilled models, and Models suffixed with “-L” are LLaMA-distilled models.}
    \label{tbl:performance_models}
\end{table*}

\begin{table*}[t]
    \centering
    \resizebox{\textwidth}{!}{
    \begin{tabular}{c||c|c|c|c|c|c|c|c|c}
        \hline
         TELeR Level
         & \begin{tabular}[c]{@{}c@{}} Anchoring\end{tabular}
         & \begin{tabular}[c]{@{}c@{}} Availability\end{tabular}
         & \begin{tabular}[c]{@{}c@{}} Confirmation\end{tabular}
         & \begin{tabular}[c]{@{}c@{}} Framing\end{tabular}
         & \begin{tabular}[c]{@{}c@{}} Overattribution\end{tabular}
         & \begin{tabular}[c]{@{}c@{}} Prospect Theory\end{tabular}
         & \begin{tabular}[c]{@{}c@{}} Representativeness\end{tabular}
         & \begin{tabular}[c]{@{}c@{}} Interpretation\end{tabular}
         & \begin{tabular}[c]{@{}c@{}} Average\end{tabular}\\
         \hline\hline
TELeR-L1		& 0.611	& 0.722	& 0.870	& 0.473	& \highlight\textbf{0.667}	& 0.542	& 0.442	& 0.555	& 0.610	\\
\hline
TELeR-L2		& \highlight\textbf{0.649}	& 0.821	& \highlight\textbf{0.896}	& 0.509	& 0.585	& 0.590	& 0.396	& 0.611	& 0.632	\\
\hline
TELeR-L3		& 0.625	& 0.868	& 0.874	& \highlight\textbf{0.561}	& 0.523	& 0.599	& 0.502	& 0.605	& 0.644	\\
\hline
TELeR-L4		& 0.621	& \highlight\textbf{0.871}	& 0.879	& 0.538	& 0.547	& 0.625	& \highlight\textbf{0.507}	& \highlight\textbf{0.651}	& 0.655	\\
\hline
TELeR-L5		& 0.643	& 0.866	& 0.891	& 0.549	& 0.539	& \highlight\textbf{0.630}	& 0.485	& 0.640	& \highlight\textbf{0.655}	\\
\hline
    \end{tabular}
    }
    \caption{Average bias resistance scores for each TELeR level. Higher ($\uparrow$) scores indicate resistance, and lower ($\downarrow$) scores indicate susceptibility. Top performers are highlighted and marked in \textbf{bold}.}
    \label{tbl:performance_teler}
\end{table*}





To visualize these patterns, we plot the susceptibility scores as heatmaps, as shown in Figure \ref{fig:heatmap_score}. From these, several key observations emerge:

\begin{enumerate}\setlength\itemsep{0cm}
    \item \textbf{Bias-specific variance}: Some biases, such as \textit{Representativeness} and \textit{Prospect Theory}, induce higher error rates across models, while others are more easily resisted.
    \item \textbf{Prompt sensitivity}: Prompt detail has a non-uniform effect across biases; \textit{Availability} shows substantial improvement with added context (by up to \textbf{14.9\%}), while \textit{Overattribution} shows degradation (by up to \textbf{8.8\%}).
    \item \textbf{Architectural robustness}: Models trained with explicit reasoning objectives (e.g., \textit{QwQ-32B}, \textit{DeepSeek-R1}) demonstrate improved resistance relative to similarly-sized peers.
    \item \textbf{Scaling Effects}: Larger models (\textit{o4-mini}, \textit{GPT-4.1}, and \textit{QwQ-32B}) generally exhibit lower bias, suggesting that larger models possess more bias resistance.
\end{enumerate}

\subsection{Observable Trends}
\subsubsection{Bias-Specific Variance}
The heatmaps in Figure~\ref{fig:heatmap_score} are ordered horizontally by average resistance, revealing a wide variance across biases. Models show relatively strong resistance to \textit{Confirmation} and \textit{Availability} biases, while \textit{Representativeness}, \textit{Framing}, and \textit{Prospect Theory} induce consistently higher error rates. These trends hold across architectures and prompt levels, suggesting that some cognitive heuristics are more naturally aligned with the inductive biases of LLMs. In particular, biases involving probabilistic reasoning (e.g., Representativeness) may be harder to override, as they depend more on implicit distributional knowledge than surface-level associations.

\subsubsection{Prompt Sensitivity}
Across most biases, susceptibility decreases with increasing TELeR level (Table \ref{tbl:performance_teler}), consistent with the idea that more contextual grounding supports more accurate reasoning. For example, \textit{Availability} bias sees a marked drop at Level 2, likely due to early contextual disambiguation. However, the trend is not uniform: \textit{Overattribution} becomes more prominent at Level 3, possibly due to the inclusion of spurious or overly specific details in the prompt. These findings suggest that prompt engineering can mitigate or amplify bias, depending on the type of bias and degree of contextual detail, as explored further in Section~\ref{ssec:effect_of_prompt_detail}.

\subsubsection{Architectural Robustness}
Reasoning-oriented models such as \textit{QwQ-32B} and \textit{DeepSeek-R1} show increased resistance to \textit{Framing} bias, even when size-matched against less capable peers (Figure \ref{fig:heatmap_score}). This suggests that architectural or training-time interventions may reduce certain biases. Notably, this effect is limited: we do not observe comparable gains for other biases, indicating that Framing bias uniquely benefits from intermediate reasoning steps.



\subsubsection{Scaling Effects}
Larger models like \textit{o4-mini}, \textit{GPT-4.1}, and \textit{QwQ-32B} consistently perform highly across all biases. In general, bias susceptibility appears to decrease with model size up to a point (Figure \ref{fig:heatmap_score}), after which performance seems to level off. This suggests possible diminishing returns from scale alone, which indicates that excessively large parameter counts may not be necessary when bias is a concern.

\subsection{Case Studies}
To illustrate how these biases manifest in practice, we present two example responses: one which demonstrates an unbiased answer to a bias with high average resistance (Availability bias), and one which demonstrates a biased answer to a bias with low average resistance (Representativeness bias).

The first example is taken as a response to a prompt targeting Availability bias. The response was generated by Gemma 2-9B at TELeR level 2, with an inference temperature of 0.2. The model was asked to determine whether or not the user should fly somewhere after they had recently heard about a plane crash, and it answered with:

\begin{quote}\emph{
    **B. Yes, because the likelihood of a plane crash is actually very low** Here's why:* **Anecdotal Evidence:**  Stories about plane crashes, while impactful, are anecdotal evidence. They don't represent the overall statistical reality. * **Actual Likelihood:**  Flying is statistically one of the safest modes of transportation. The chances of being in a fatal plane crash are extremely low. It's natural to be concerned about safety, but it's important to base decisions on facts and statistics rather than isolated incidents.
}\end{quote}

This response reflects an unbiased, rational judgment of the situation. Despite the emotional weight placed on having recently been exposed to a plane crash, the model bases its judgment on proper statistical reasoning to recommend that the user would likely be fine to take the flight.

In contrast, the second example is a response taken from a prompt targeting Representativeness bias, generated by the same model under the same configuration.

The model was asked to judge who was more likely to be a photographer: the casually-dressed Benjamin, the formally-dressed James, or either one. The model responded with:

\begin{quote}\emph{
    **A. Benjamin** While there's no definitive link between clothing style and profession, the description of Benjamin's attire (baggy designer shirt, tattoos)  is more commonly associated with creative fields, which could align with a photographer.
}\end{quote}

Though the model recognizes the potential for bias in the question via stereotyping, it ultimately provides a definite, biased answer aligned with that stereotyping, rather than expressing uncertainty and selecting the unbiased answer: "It could be either". In doing so, the model perfectly mirrors the kind of irrational but intuitive judgment that humans can make when affected by the Representativeness bias.

\section{Analysis}

While these trends suggest meaningful relationships between model configurations and bias susceptibility, they are not yet conclusive. To examine these effects more rigorously, we conduct a targeted analysis to explore how key factors—such as model size, TELeR level, and bias type—interact to influence bias resistance across models.
Detailed statistical significance tests supporting this analysis are included in Appendix \ref{sec:appendix-significance}

\begin{figure}[ht]
    \centering
    \includegraphics[scale = 0.465]{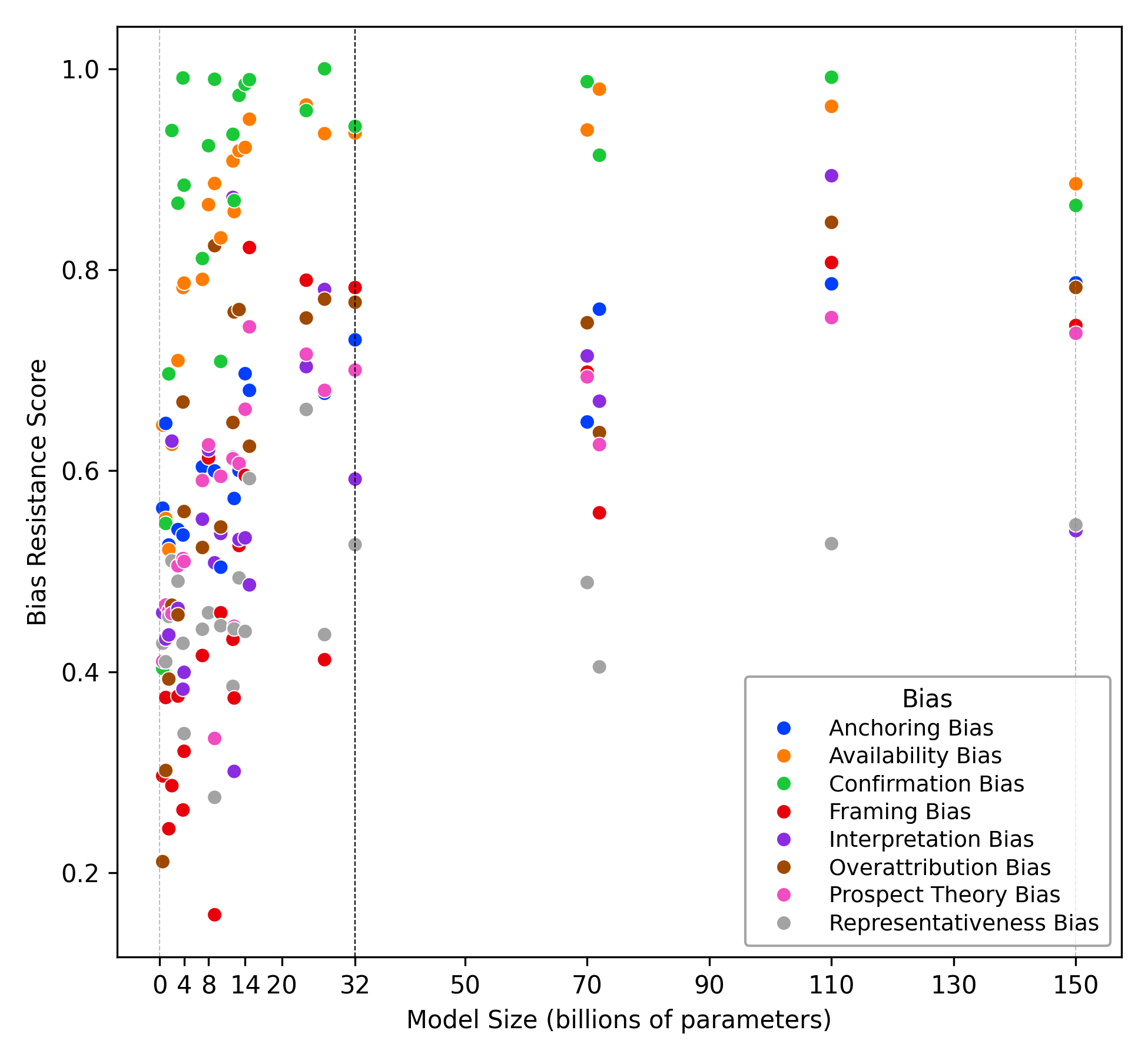}
    \vspace{-8mm}
    \caption{Scatter plot showing changes in average bias resistance scores across model sizes. Dot colors indicate different biases.}
    \label{fig:size_comparison}
\end{figure}
\vspace{-2pt}

\subsection{Effect of Model Size}
One of the most salient factors in task-oriented models is their \textbf{size}. As shown in Figure \ref{fig:size_comparison}, larger models generally exhibit higher resistance to bias, with resistance increasing as parameter counts grow. However, this effect diminishes as models scale past 32B parameters, with the most significant gains in bias resistance occurring between 0.5B and 32B parameters. On average, models see \textbf{56\%} of their improvement in this range, suggesting that future gains will likely depend more on training strategy or architecture than on sheer size.



\begin{figure}[ht]
    \centering
    \includegraphics[scale = 0.465]{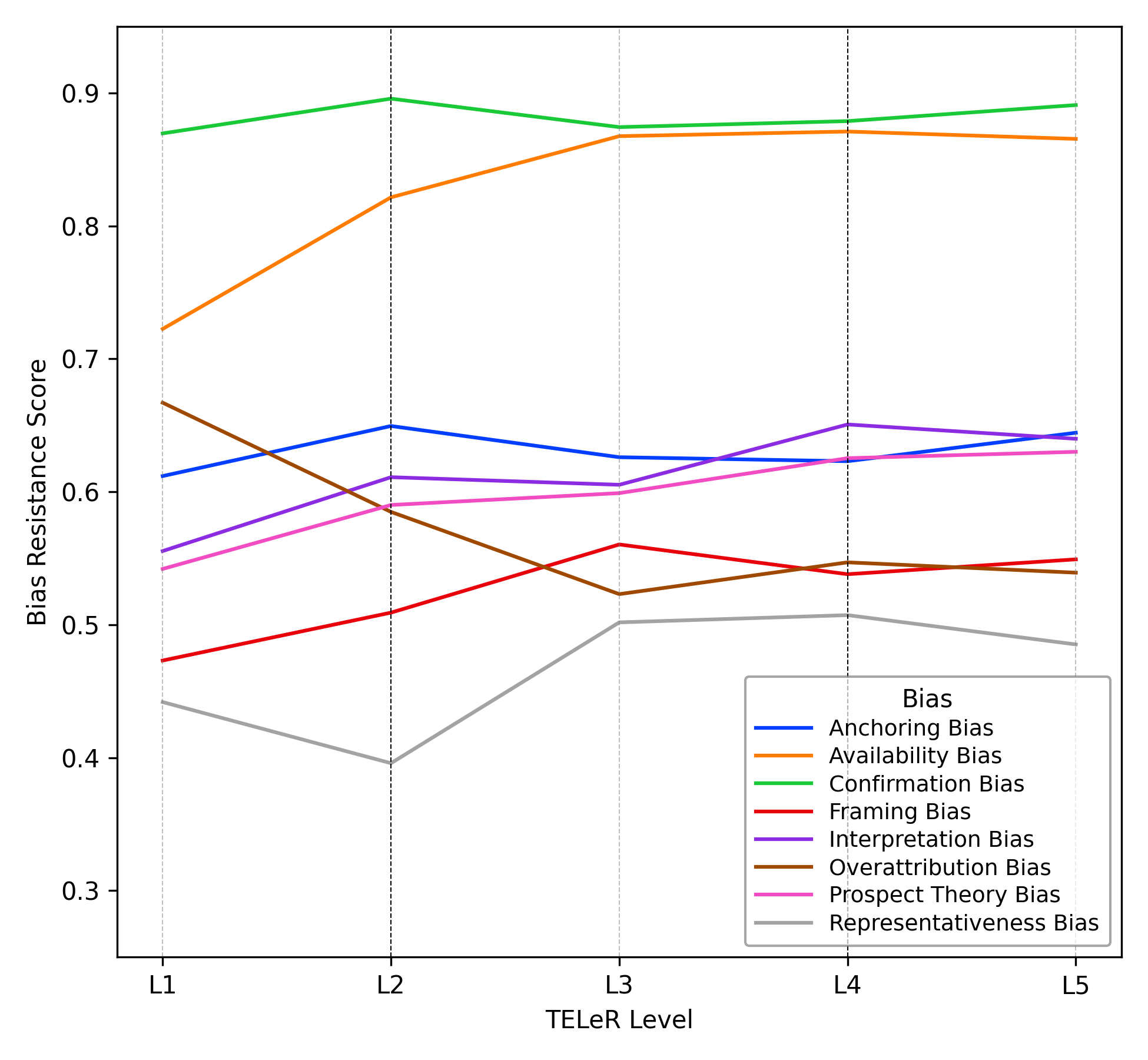}
    \vspace{-8mm}
    \caption{Line graph showing changes in average bias resistance scores across TELeR prompting levels. Line colors indicate different biases.}
    \label{fig:teler_comparison}
\end{figure}
\vspace{-2pt}

\subsection{Effect of Prompt Detail}\label{ssec:effect_of_prompt_detail}
To further understand how varied prompting styles affect bias resistance, we compared model performance across different TELeR levels for each bias. The results depicted in Figure \ref{fig:teler_comparison} show that certain levels of detail in the prompt are more influential in shaping model responses than others, with the effects varying by bias. From this, we observe three distinct response patterns to higher prompt detail:
\vspace{-2mm}
\subsubsection{Level 2: Clear Directives}
The first noticeable pattern emerges at \textbf{TELeR Level 2}, where the prompt provides basic situational context and clearly restates the high-level directive. Models tend to respond positively to this additional context, and resistance to \emph{Anchoring} and \emph{Confirmation} bias peaks at this level (Figure \ref{fig:teler_comparison}). For \emph{Anchoring}, the added contextual grounding helps distract from the unrealistic anchor, and for \emph{Confirmation}, the clear directive encourages more critical evaluation, reducing the tendency to conform to biased, agreeable responses. We also see improvements in resistance to \emph{Availability}, \emph{Framing}, \emph{Interpretation}, and \emph{Prospect Theory} biases by \textbf{3.6-9.9\%}, making clear directives a valuable strategy for reducing a broad range of biases.

\subsubsection{Levels 3-4: Appeals to Rationality}
A second pattern appears at \textbf{TELeR Levels 3 and 4}, which guide the model's output through specific subtask enumerations (Level 3) and clear descriptions of desirable response qualities (Level 4). These levels often explicitly instruct the model to consider each option carefully and prioritize rationality, using phrases like ``objective'', ``accurate'', and ``fair''. These targeted prompts emphasizing logical reasoning consistently improve the model's ability to resist biases by \textbf{6.5-14.9\%} (Figure \ref{fig:teler_comparison}).

\subsubsection{Harmful Details}
In contrast, certain higher levels of detail can inadvertently \emph{increase} bias susceptibility. \emph{Overattribution} bias, in particular, demonstrates a significant \textbf{8.2\%} drop in resistance when moving from Level 1 to Level 2, suggesting that the additional contextual details inadvertently focus the models' attention on the bias-eliciting portions of the prompt. Similarly, \emph{Representativeness} sees a \textbf{4.6\%} decrease in resistance at Level 2, likely due to an overemphasis on biased, stereotypical cues in the prompt (Figure \ref{fig:teler_comparison}). Prompts that aim to reduce these biases should focus on neutral and objective language, avoiding over-emphasis of potentially biased elements.
\section{Conclusions \& Future Work}\label{sec:conclusion}


In this work, we present a comprehensive evaluation of eight targeted cognitive biases across a diverse set of \textbf{45} LLMs, analyzing over \textbf{2.8 million} responses generated through controlled prompt variations. Our work makes several key contributions: we introduce a novel framework for evaluating LLMs based on a multiple-choice question-answering task; curate a new dataset of $220$ psychologist-crafted judgment and decision scenarios spanning eight cognitive biases; and propose a scalable approach to automatically generate unique LLM prompts from a small set of expert-written templates; and finally, we conduct extensive experiments to explore the factors influencing a model's susceptibility to each bias.



Our experiments demonstrate that \emph{all} LLMs exhibit susceptibility to cognitive biases, with susceptibility rates ranging from \textbf{17.8\%} to \textbf{57.3\%} on average across the models and biases tested. Further analysis reveals that \emph{larger models} and \emph{increased prompt detail} tend to reduce bias susceptibility. Specifically, models larger than 32B parameters can reduce bias susceptibility by up to \textbf{39.5\%}, and prompts with greater detail (TELeR Levels 3-5) improve bias resistance by up to \textbf{14.9\%} in most cases. These findings highlight that a combination of model size and thoughtful prompting strategies can mitigate bias, from which we derive the following best practices for interacting with LLMs:

\begin{enumerate}\setlength\itemsep{0cm}
    \item \textbf{Include more details when prompting.} \\
    In most cases, adding situational context or descriptive information can help reduce bias.
    \item \textbf{Avoid ambiguous language.} \\
    Instead of: \textit{I just heard X, \textbf{could} that be true?} \\
    Use: \textit{I just heard X, \textbf{is} that true?}
    \item \textbf{Explicitly request rational responses.} \\
    When asking: \textit{What's a normal amount for X?} \\
    Add: \textit{Please give an accurate estimate.}
    \item \textbf{Avoid leading or biased language.} \\
    Instead of: \textit{Do I need more than X of Y?} \\
    Use: \textit{How much Y should I get?}
    \item \textbf{Use an appropriately-sized model.} \\
    Larger models are generally less susceptible to bias. Notably, models larger than 32 billion parameters see diminished benefit.
\end{enumerate}

Future work may additionally leverage the groundwork established here to
extend this evaluation pipeline to other domains beyond decision-making and cognitive bias, such as ethical reasoning, fairness, or task-oriented contexts. By adapting this framework to different settings, it may offer valuable insights into how LLMs perform on a variety of cognitive tasks, revealing patterns in both their biases and decision-making strategies.
\section{Limitations}\label{sec:limitations}

We delineated a subset of eight cognitive biases to provide a structured and holistic initial evaluation of bias in LLMs. While this scope was limited intentionally, evaluating additional biases, particularly those that are domain-specific, could strengthen the robustness of the trends observed.


Though we tested 45 different models across different architectures and sizes, new models are created rapidly and constantly upset the status quo. As such, following up by testing more, future models would again only help to strengthen the robustness of the conclusions drawn from this work. 

We explicitly avoid the use of LLMs in augmenting our scenario templates for testing. Though this helps to preserve the validity of the study, it also means that we employ a more straightforward method to augment the set of human-authored scenarios into a sufficient number of test prompts. Future improvements to this LLM-free augmentation pipeline could increase the speed at which datasets can be generated, as well as alleviate the burden placed on experts of a subject when generating scenario templates.

Finally, cognitive biases only represent one facet of potential irrationality in both humans and LLMs. Though some work already exists, future analyses could be expanded to include ethical, social, or cultural biases that would provide a more comprehensive understanding of how LLMs exhibit bias under varied user prompting styles.

\bibliography{anthology}
\bibliographystyle{acl_natbib} 

\appendix
\section{Appendix}

\subsection{Algorithms}


Algorithms \ref{alg:template_filler} and \ref{alg:answer_extractor} describe the procedures used by the \emph{Template Filler} and \emph{Answer Extractor} modules, as described in Sections \ref{subsec:generation} and \ref{subsec:extraction}. The \emph{Answer Extractor} uses Levenshtein distance \cite{Levenshtein1965-gk} for fuzzy string matching when computing the \textbf{presence} score described in Section \ref{subsec:extraction}.

\begin{algorithm}
\caption{Template Filler}\label{alg:template_filler}
\vskip.2\baselineskip
\textbf{Input:} Collection of templates $T$\\
\textbf{Output:} Collection of prompts $P$
\par\vskip.3\baselineskip\hrule height .5pt\par\vskip.3\baselineskip
\begin{algorithmic}
\For{$t$ in $T$}
    \For{$t_i$ in $t$}
        \If{$t_i$ is numeric}
            \State $t_i \to expression$
            \State $sample \gets uniform(min, max)$
            \State $t_i \gets eval(expression, sample)$
        \ElsIf{$t_i$ is not numeric}
            \State $t_i \to phrase\_list$
            \State $t_i \gets random(phrase\_list)$
        \EndIf
    \EndFor
\EndFor
\end{algorithmic}
\end{algorithm}

\begin{algorithm}[ht]\small
\caption{Answer Extractor}\label{alg:answer_extractor}
\vskip.2\baselineskip
\textbf{Input:} Collection of answers $A$, LLM response $r$, Positive terms $P_{pos}$, Negative terms $N_{pos}$, Similarity threshold $t$\\
\textbf{Output:} Collection of answer scores $S_a$
\par\vskip.3\baselineskip\hrule height .5pt\par\vskip.3\baselineskip
\begin{algorithmic}
\For{$a$ in $A$}
        \State $max\_score \gets 0$
        \State $score\_sum \gets 0$
        \State $M \gets []$
        \For{$i = 0$ to $|r| - |a|$}
            \State $w_i \gets r[i : i+|a|]$
            \State $score \gets fuzzy\_match(w_i, a)$
            \State $score\_sum \gets score\_sum + score$
            \If{$score$ > $max\_score$}
                \State $max\_score \gets score$
            \EndIf
            \If{$score$ > $t$}
                \State Append $w_i$ to $M$
            \EndIf
        \EndFor
        \State $avg\_score \gets score\_sum / (|r| - |a|)$
        \State $score_p \gets (avg\_score * .25 + max\_score * .75)$
        \State $weight_s \gets calc\_C_m(M, P_{pos}, N_{pos})$
        \State Append $(score_p * weight_s)$ to $S_a$
\EndFor
\end{algorithmic}
\end{algorithm}

\subsection{Statistical Significance Tests}\label{sec:appendix-significance}





\begin{table}[!hb]
    \centering
    \resizebox{\columnwidth}{!}{
    \begin{tabular}{c||c|c|c}
        \hline
         Bias
         & \begin{tabular}[c]{@{}c@{}} F-value\end{tabular}
         & \begin{tabular}[c]{@{}c@{}} $p$\end{tabular}
         & \begin{tabular}[c]{@{}c@{}} $\beta$\end{tabular}\\
         \hline\hline
         Anchoring        & 0.001 & 0.982 & -0.001     \\
         \hline
         Availability        & 0.292 & 0.589 & 0.014     \\
         \hline
         Confirmation        & 10.861        & 0.001 & -0.058    \\
         \hline
         Framing        & 0.066 & 0.798 & 0.005     \\
         \hline
         Interpretation        & 0.070 & 0.791 & -0.006    \\
         \hline
         Overattribution        & 2.698 & 0.101 & -0.046     \\
         \hline
         Prospect Theory        & 0.071 & 0.790 & -0.008     \\
         \hline
         Representativeness        & 0.221 & 0.638 & -0.014    \\
         \hline
    \end{tabular}
    }
    \vspace{-2mm}
    \caption{ANOVA results assessing the effect of inference temperature on bias resistance scores for each bias, controlled for model identity.}
    \label{tbl:question_3}
\end{table} 
\begin{table}[]
    \centering
    \label{tbl:question_4_part_2}
    \resizebox{\columnwidth}{!}{
    \begin{tabular}{c||c|c|c}
        \hline
         Bias
         & \begin{tabular}[c]{@{}c@{}} F-value\end{tabular}
         & \begin{tabular}[c]{@{}c@{}} $p$\end{tabular}
         & \begin{tabular}[c]{@{}c@{}} $R^2$\end{tabular}\\
         \hline\hline
        Anchoring       & 54.464        & < 0.01        & 0.021 \\
        \hline
        Availability    & 139.879       & < 0.01        & 0.087 \\
        \hline
        Confirmation    & 89.740        & < 0.01        & 0.057 \\
        \hline
        Framing & 260.655       & < 0.01        & 0.119 \\
        \hline
        Interpretation  & 124.266       & < 0.01        & \textbf{0.201} \\
        \hline
        Overattribution     & 42.583        & < 0.01        & 0.080 \\
        \hline
        Prospect        & 78.631        & < 0.01        & 0.050 \\
        \hline
        Representativeness      & 5.621 & 0.018 & < 0.01        \\
        \hline
    \end{tabular}
    }
    \vspace{-2mm}
    \caption{Ordinary least squares (OLS) regression results assessing the effect of model size on bias resistance scores for each bias. Bold $R^2$ values indicate a substantial proportion of variance explained ($R^2 > 0.2$). Values less than 0.01 are truncated for clarity.}
\end{table} 
\begin{table}[]
    \centering
    \label{tbl:question_4_part_3}
    \resizebox{\columnwidth}{!}{
    \begin{tabular}{c||c|c|c}
        \hline
         Bias
         & \begin{tabular}[c]{@{}c@{}} F-value\end{tabular}
         & \begin{tabular}[c]{@{}c@{}} $p$\end{tabular}
         & \begin{tabular}[c]{@{}c@{}} $R^2$\end{tabular}\\
         \hline\hline
        Anchoring       & 0.085 & 0.771 & < 0.01        \\
        \hline
        Availability    & 19.710        & < 0.01        & 0.013 \\
        \hline
        Confirmation    & < 0.01        & 0.986 & < 0.01        \\
        \hline
        Framing & 530.211       & < 0.01        & \textbf{0.215} \\
        \hline
        Interpretation  & 18.109        & < 0.01        & 0.035 \\
        \hline
        Overattribution     & 1.234 & 0.267 & < 0.01        \\
        \hline
        Prospect        & 58.239        & < 0.01        & 0.038 \\
        \hline
        Representativeness      & 0.054 & 0.816 & < 0.01        \\
        \hline
    \end{tabular}
    }
    \vspace{-2mm}
    \caption{Ordinary least squares (OLS) regression results assessing the effect of reasoning capabilities on bias resistance scores for each bias. Bold $R^2$ values indicate a substantial proportion of variance explained ($R^2 > 0.2$). Values less than 0.01 are truncated for clarity.}
\end{table} 
\begin{table}[]
    \centering
    \label{tbl:validity}
    \begin{tabular}{c||c|c|c}
        \hline
         Bias
         & \begin{tabular}[c]{@{}c@{}} F-value\end{tabular}
         & \begin{tabular}[c]{@{}c@{}} $p$\end{tabular}
         & \begin{tabular}[c]{@{}c@{}} $R^2$\end{tabular}\\
         \hline\hline
         Combined        & 26.6 & < 0.01 & 0.430    \\
         \hline
    \end{tabular}
    \caption{The discriminant validity interaction effect between evaluation prompt sets and model identity. Values less than 0.01 are truncated for clarity.}
\end{table}

\end{document}